\documentclass[11pt,a4paper]{article}
\usepackage[hyperref]{blanc}
\usepackage{times}
\usepackage{latexsym}
\usepackage{graphicx} % added
\usepackage{booktabs} % added
\usepackage{placeins} % added
\usepackage{microtype} % added
\usepackage{enumitem} % added
\usepackage{microtype}

\aclfinalcopy % Uncomment this line for the final submission

\title{Fill in the BLANC:\\
Human-free quality estimation\\
of document summaries}

\author{Oleg Vasilyev, Vedant Dharnidharka, John Bohannon \\
  Primer Technologies Inc. \\
  San Francisco, California \\
  \texttt{{oleg,vedant,john}@primer.ai}\\}

\date{}

\begin{document}
\maketitle
\begin{abstract}
We present BLANC, a new approach to the automatic estimation of document summary quality. Our goal is to measure the functional performance of a summary with an objective, reproducible, and fully automated method. Our approach achieves this by measuring the performance boost gained by a pre-trained language model with access to a document summary while carrying out its language understanding task on the document's text. We present evidence that BLANC scores have as good correlation with human evaluations as do the ROUGE family of summary quality measurements. And unlike ROUGE, the BLANC method does not require human-written reference summaries, allowing for fully human-free summary quality estimation.
\end{abstract}

\section{Introduction}
Two most widely used methods for measuring the quality of a summary are ROUGE \cite{OriginalROUGE} and human evaluation \cite{Wojciech2019Neural}.

The ROUGE family of methods are well-defined and reproducible. However, these methods typically require a human-written reference summaries for comparison, completely disregarding the original document text. Even if one assumes that a reference summary is available and of optimal quality, the ROUGE method is limited to measuring a mechanical overlap of text tokens with little regard to semantics. This deficiency may be partially addressable through measurement of the similarity not of text tokens but named entities or other preprocessed features \cite{Yuning2019Facet, Arman2016Revisiting, Fatma2015Keyphrase, Jun2015Better, Kavita2018ROUGE2} or embeddings \cite{Wei2019MoverScore, Tianyi2020BERTScore, Yang2020Supert}. In the latter work \cite{ Yang2020Supert} the references are not human-written but unsupervisedly constructed from selected salient sentences. An overlap can be measured as well between summary and document text \cite{Liqun2017Efficient}.

Human evaluation of summary quality is far more meaningful and powerful than ROUGE, but it is far less reproducible. Summary quality estimation is a cognitively demanding and highly subjective task. Humans are also vulnerable to biases, such as the preference for phrases and sentences copied directly from the document text into summaries \cite{Daniel2020FineTuning}. Improving human evaluation may require prompting labelers to pay higher attention \cite{Hardy2019HighRES}, as well as splitting quality scores into multiple dimensions such as fluency, informativeness, and factual correctness \cite{Wojciech2019Neural, Wojciech2017Evaluating, Lisa2018Robust}. Even if humans can be trained to be more reliable, reproducible estimators of summary quality, they will forever remain a slow, expensive, limiting resource.

One possible route to a better automatic method for summary quality estimation is to train a model on document summaries annotated with  human quality scores \cite{Louis2009Automatically, Louis2013Automatically, Stratos2019Sum}. Such a model could be used to evaluate  summaries without further human involvement. But even if such a model could achieve high agreement with human labelers, its performance would only be as objective and reproducible as the summary quality scores generated by one particular group of humans on a particular group of documents. Such a model may not generalize beyond the domain and style of the training samples unless they are a massive, representative sample of all documents of interest.

A more fundamental approach to the problem is to estimate how "helpful" a summary is for the task of understanding a text. For example this might be achieved through a series of question-answers \cite{Matan2019Question, Ping2018SemanticQA, Thomas2015Answers}. However, with this approach one must choose from a vast set of questions one might ask of a text, presupposing knowledge of the document itself and seriously limiting its reproducibility.

In the following section we suggest a new approach that is fundamentally justifiable as an estimator of summary quality, as well as being conceptually simple and reproducible.

\section{Methods}
\subsection{Introducing BLANC}
An ideal estimator should directly test how helpful a summary is to its readers. It should reliably estimate quality across a broad range of document domains and styles. And yet it should achieve this without requiring ornate preconditions and presuppositions about the text being summarized. If this estimator relies upon an existing base model, that model should be well-documented, well-understood and widely used.

We propose BLANC\footnote{According to ancient tradition, we should adorn our newly created jargon term with a bacronymic justification. The term BLANC is a nod to its proud lineage of French color words that began with the BLEU method for evaluating machine translation and ROUGE for summarization. BLANC is also a reference to the method's core task of "filling in the blanks" in the masked token task. But to honor tradition we offer this: Bacronymic Language model Approach for summary quality estimatioN. Cool?} as a replacement for the ROUGE family of summary quality estimators. 

We define BLANC as a measure of how well a summary helps an independent, pre-trained language model while it performs its language understanding task on a document. We focus on the masked token task, also known as the Cloze task \cite{Wilson1953Cloze}, in which a model is challenged to reconstruct obscured spans of text. We use the well-known BERT language model \cite{Jacob2018BERT} pre-trained to predict masked text tokens (words or sub-words). The BERT tokenizer represents the majority of the most frequently used words as single tokens, while splitting less common words into two or more. 

We present two versions of BLANC, which we dub BLANC-help and a BLANC-tune. These measures are described in detail in the following sections. The essential difference between them:
\begin{enumerate}[topsep=0pt,itemsep=-1ex,partopsep=1ex,parsep=1ex]
    \item BLANC-help uses the summary text by directly concatenating it to each document sentence during inference.
    \item BLANC-tune uses the summary text to fine-tune the language model, and then processes the entire document.
\end{enumerate}
Thus with BLANC-help, the language model refers to the summary each time it attempts to understand a part of the document text. While with BLANC-tune, the model learns from the summary first, and then uses its gained skill to help it understand the entire document.

\subsection{BLANC-help}
\begin{figure}[htb]
    \centering
    \includegraphics[width=7.9cm]{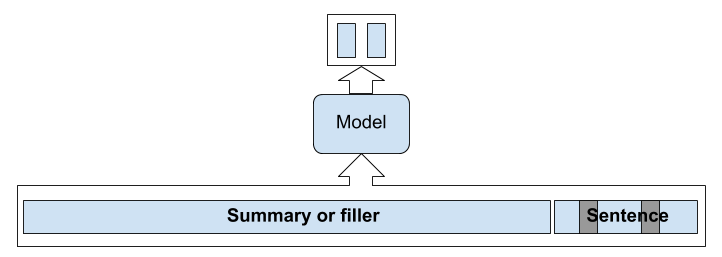}
    \caption{BLANC-help of summary quality is defined by the difference in accuracy of two reconstructions of masked tokens: with summary vs. filler concatenated in front of the sentence with masked tokens. The model input is a summary (or filler) + sentence with masked (grey) tokens. The output is the unmasked tokens.}
    \label{fig:Measure_help}
\end{figure}

The algorithm for obtaining BLANC-help scores is illustrated in Figure \ref{fig:Measure_help}.

There are many possible choices for how to mask the tokens. Our aim is to evenly cover all tokens in a sentence with a certain frequency, for the sake of full reproducibility. A random coverage is also possible, but it is not as conveniently reproducible.

The unmasking is done twice for each sentence of the text and for each allowed choice of masked tokens in the sentence. First, the unmasking is done for input composed of the summary concatenated with the sentence. Second, the unmasking is done for input composed of a "filler" concatenated with the sentence. The filler has exactly the same lengths as the summary, but each summary token is replaced by a period symbol ("."). After iterating over all sentences and over all the allowed choices of masking, we end up with four total counts of successful and unsuccessful unmasking $S_{ij}, i=0,1; j=0,1$. Here the index $i$ equals 0 or 1 - for unsuccessful (0) or successful (1) unmasking for the filler-input. The index $j$ is defined the same way for the summary-input. For example, $S_{01}$ is the count of cases where the filler-input was unsuccessful and the summary-input was successful.

We define BLANC-help as the difference between the accuracy $A_s$ of unmasking with the summary and the accuracy $A_f$ of unmasking with the filler:

%\[BLANC_{\text{help}} = A_s - A_f = \frac{S_{01} - S_{10}}{S_{total}}\]
\[BLANC_{help} = A_s - A_f = \frac{S_{01} - S_{10}}{S_{total}}\]

The accuracies are $A_s = (S_{11} + S_{01}) / S_{total}$ and $A_f = (S_{11} + S_{10}) / S_{total}$. The total count is $S_{total} = S_{00} + S_{11} + S_{01} + S_{10}$. The BLANC value can range from -1 to 1, but as shown in next sections the typical values are between 0 (summary is useless) and 0.3 (summary provides 30\% help).

The algorithm for BLANC-help is shown in more detail in Figure \ref{fig:Algo_help}.    
 \begin{figure}
     \centering
     \begin{tabular}{l}
     \toprule
     Given: $summary$; $text$; $model$;\\ 
     \hspace{.4cm}Parameters $M=6$, $L_{min}=4$\\  \\
     
     Initialise $filler = "." * length(summary)$\\
     Initialise $S_{00}$, $S_{01}$, $S_{10}$, $S_{11}$ to zero\\
     \bf{for} $sentence$ in $text$:\\
     \hspace{.4cm}\bf{for} $i_{0}$ in range from $1$ to $M$:\\
     \hspace{.4cm}\hspace{.4cm}Mask each $i$th word if $(i-i_{0})\%M == 0$\\ 
     \hspace{.4cm}\hspace{.4cm}\hspace{.4cm}and if $length(word)>= L_{min}$\\
     \hspace{.4cm}\hspace{.4cm}$input_{base} = filler + sentence$\\
     \hspace{.4cm}\hspace{.4cm}$input_{help} = summary + sentence$\\
     \hspace{.4cm}\hspace{.4cm}$out_{base} = model(input_{base})$\\
     \hspace{.4cm}\hspace{.4cm}$out_{help} = model(input_{help})$\\
     \hspace{.4cm}\hspace{.4cm}{\bf{for each}} position $i$ in masked tokens:\\
     \hspace{.4cm}\hspace{.4cm}\hspace{.4cm}$k = int(out_{base}[i] == sentence[i])$\\
     \hspace{.4cm}\hspace{.4cm}\hspace{.4cm}$m = int(out_{help}[i] == sentence[i])$\\
     \hspace{.4cm}\hspace{.4cm}\hspace{.4cm}$S_{km} += 1$\\
     $B = (S_{01} - S_{10}) / (S_{00} + S_{11} + S_{01} + S_{10})$\\
     \bottomrule
     \end{tabular}
     \caption{BLANC-help $B$ for quality of summary.}
     \label{fig:Algo_help}
 \end{figure}

Since the BERT model deals with tokens rather than words, we can choose to mask tokens rather than words. In typical news documents only about 10\% of words are split by the BERT tokenizer into two or more tokens. Such "composite" words (not existing in the BERT vocabulary) should be particularly valuable in estimating the helpfulness of a summary. In a version dealing with tokens rather than words it is natural to always allow masking of composite words regardless of their length.

The setting $L_{min}=4$ allows the masking only of sufficiently long words (4 or more characters), because shorter words are typically easier to predict, with or without the help of a summary. 

The value $M = 6$ in Figure \ref{fig:Algo_help} is a natural choice because the standard BERT model is trained by masking 15\% of tokens, which makes about one-sixth of tokens eligible to be masked.  

We found that altering the filler has a negligible effect on the measures. The reason we use the filler is to avoid any effect of the length of input on the action of the model.

\subsection{BLANC-tune}
\begin{figure}[!htbp]
    \centering
    \includegraphics[width=7.9cm]{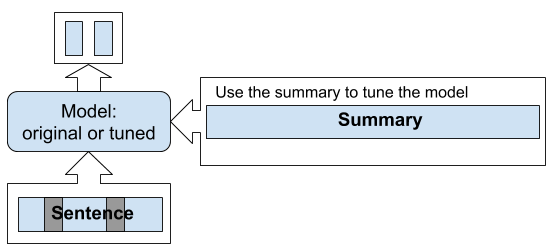}
    \caption{BLANC-tune of summary quality is defined by the difference in accuracy of two reconstructions of masked tokens: with model tuned on the summary vs. with the original model. Both models are given the same input: a sentence with masked (grey) tokens. Each model outputs the unmasked tokens.}
    \label{fig:Measure_tune}
\end{figure}

The algorithm for obtaining BLANC-tune is illustrated in Figure \ref{fig:Measure_tune}.

For calculating this measure, the model first learns from the summary, and then we observe how helpful this learning was in reconstructing masked tokens in text sentences.

As in the case of BLANC-help, we define BLANC-tune by comparing the accuracy of two reconstructions: one that does use the summary, and another that does not. In the case of BLANC-help, this was the difference between placing the summary vs. placing the filler in front of a sentence. Now, in the case of BLANC-tune, we compare the performance of a model fine-tuned on the summary text vs. a model that has never seen the summary.

The task, using a model to unmask tokens, is performed the same way as for BLANC-help, except that the input is simply a document sentence with masked tokens.

The tuning of the model is done on an extremely small dataset (derived from the summary text), in which each sample is the very same summary but with different tokens masked. The masking in the summary is done according to the original BERT pre-training strategy. Unmasking must be performed for 15\% randomly selected tokens, of which 80\% are masked, 10\% are replaced by random tokens, and 10\% are left unchanged. To ensure coverage of tokens, we select and shuffle all eligible tokens, and then go through them to generate samples for the BLANC-tune dataset.

The algorithm for BLANC-tune is shown in more detail in Figure \ref{fig:Algo_tune}.

 \begin{figure}[htb]
     \centering
     \begin{tabular}{l}
     \toprule
     Given:\\ 
     \hspace{.4cm}$summary$; $text$; $model$;\\
     \hspace{.4cm}probability $p_{mask}=0.15$ of masking tuning;\\ 
     \hspace{.4cm}min length of word to be masked $L_{min}=4$;\\
     \hspace{.4cm}number of tuning passes $N=10$\\ \\
     
     \it{\# Tune the model on the summary}\\
     $N_{words} = $ number of words in summary\\
     $N_{mask} = int(N_{words} * p_{mask})$\\
     Initialize empty tuning dataset $set_{tune}$\\
     \bf{for} $i$ in range from $1$ to $N$:\\
     \hspace{.4cm}$pos = $ positions of words longer than $L_{min}$\\
     \hspace{.4cm}Random shuffle $pos$\\
     \hspace{.4cm}{\bf{until}} all position in $pos$ are used:\\
     \hspace{.4cm}\hspace{.4cm}Mask words in next $N_{mask}$ positions\\
     \hspace{.4cm}\hspace{.4cm}Add summary with masked words to $set_{tune}$\\
     Tune $model$ on $set_{tune}$. Result: $model_{tuned}$\\ \\
     
     \it{\# Compare inference with $model$ vs. $model_{tuned}$}\\
     Initialise $S_{00}$, $S_{01}$, $S_{10}$, $S_{11}$ to zero\\
     $M = integer(1 / p_{mask})$\\
     \bf{for} $sentence$ in $text$:\\
     \hspace{.4cm}\bf{for} $i_{0}$ in range from $1$ to $M$:\\
     
     \hspace{.4cm}\hspace{.4cm}Mask each $i$th word if $(i-i_{0})\%M == 0$\\ 
     \hspace{.4cm}\hspace{.4cm}\hspace{.4cm}$and$ $length(word)>= L_{min}$\\
     \hspace{.4cm}\hspace{.4cm}$out_{base} = model(sentence)$\\
     \hspace{.4cm}\hspace{.4cm}$out_{help} = model_{tuned}(sentence)$\\
     \hspace{.4cm}\hspace{.4cm}{\bf{for each}} position $i$ in masked tokens:\\
     \hspace{.4cm}\hspace{.4cm}\hspace{.4cm}$k = int(out_{base}[i] == sentence[i])$\\
     \hspace{.4cm}\hspace{.4cm}\hspace{.4cm}$m = int(out_{help}[i] == sentence[i])$\\
     \hspace{.4cm}\hspace{.4cm}\hspace{.4cm}$S_{km} += 1$\\
     $B = (S_{01} - S_{10}) / (S_{00} + S_{11} + S_{01} + S_{10})$\\
     \bottomrule
     \end{tabular}
     \caption{BLANC-tune $B$ for quality of summary}
     \label{fig:Algo_tune}
 \end{figure}

Similar to BLANC-help, there can be several variations of the measure. The details described in the previous section for BLANC-help are now applicable here in two parts of the algorithm where we must select masked tokens: for the tuning dataset, and for the inference. Any fixed version of the measure can be reproducible, with fixed seed for randomness at the tuning. In our tuning we used the same optimizer and learning rate as was used by the open source huggingface repository \cite{huggingface2019website} for training, and we found that dependency on the seed is very weak.

While BLANC-tune appears more complicated than BLANC-help, it is a promising method in that learning from a summary is separated completely from the task of understanding the document, with no concatenation required. While we use BLANC-help for the presentation of our approach in this paper, in future work we will systematically explore BLANC-tune. Our preliminary experiments showed that BLANC-tune and BLANC-help return similar values.

\subsection{Extractive summaries: no-copy-pair guard}
In the case of purely extractive summaries, the process of calculating BLANC scores may pair a summary with sentences from the text that have been copied into the summary. This exact sentence copying should be unfairly helpful in unmasking words in the original sentence. This effect may be reduced or completely eliminated by using a stronger underlying language model, especially for BLANC-tune. But a simpler solution is to include a simple guard rule into the measure: We may exclude any pairing of exact copy sentences from the calculation of the measure. In the process of iterating over text sentences, whenever a sentence contains its exact copy in the summary, it is skipped (or, alternative version, the copy is removed from the summary for this specific step in the process).

Throughout the paper we do not use the "no-copy-pair" guard, except in the corner case consideration of copying random sentences from the text, as described in the next section.

\section{Basic validation of BLANC measurement}

As part of the validation of these new measures we performed experiments to determine how a substitution of an obviously bad summary affects the measure. One example is a summary generated by selecting random words from the text. The random words summary is generated with the same length as the original summary. Our original summaries are generated for randomly selected daily news by three different methods: by Microsoft's abstractive UniML model \cite{Li2019Unified}, by semi-abstractive summarization model (based on \cite{oleg2019headline}), and by extractive LexRank model (based on \cite{Gunes2004LexRank}). The summaries generated by these models are not flawless and vary widely in overall quality when evaluated by human labelers.

In another validation experiment, we generate a "random sentences summary", which is constructed from the sentences of a document. For this example, we apply BLANC-help with the "no-copy-pair" guard introduced above. But we use the second version of the guard rule, because it is less exclusive of text sentences overall, and we also compensate for the length of the summary by replacing the copy-sentence of the summary with another sentence, rather than simply removing the copy-sentence.

\begin{figure}[htb]
    \hspace*{-0.1cm}
    \raggedleft
    \centering
    \includegraphics[width=7.9cm]{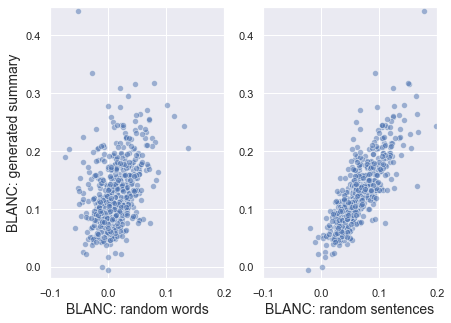}
    \caption{BLANC-help of a generated summary vs. random-words summary (left) and BLANC-help of a generated summary vs. random-sentences "summary" (right). The random-words summary is produced from random words of the same text by filling with the words the same length as the generated summary. The random-sentences summary is calculated with the no-copy-pair guard rule (version 2), but compensating for the summary length by adding more random sentences to the summary whenever needed.}
    \label{fig:random_words_sentences}
\end{figure}

BLANC-help results for both examples (in comparison to the measure of the original summaries) are shown in Figure \ref{fig:random_words_sentences}. 
We can see that the BLANC value for the real generated summary is almost always higher than the value for the random-sentences summary. This confirms that the measure takes into account the context as well as the informativeness of the summary to assess the quality.

\begin{figure}[htb]
    \centering
    \includegraphics[width=7.8cm]{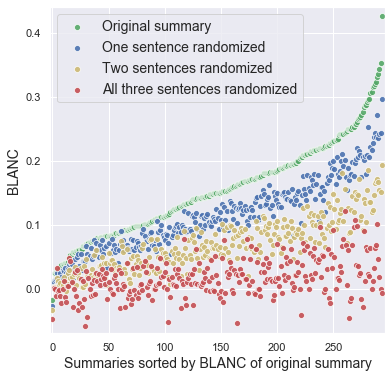}
    \caption{BLANC-help for 3-sentence summaries with one or more sentences replaced by random words from the text. The summaries are sorted by measure of the original summary.}
    \label{fig:Summary_deterioration}
\end{figure}

Selecting only summaries with exactly three sentences, we can observe how BLANC-help deteriorates if we spoil some of the sentences of the summary. We replace one, two or all three sentences with random words, keeping the same length of the resulting randomized summary as the original summary. We also take care to run on each possible choice of replacement sentences twice, and average the resulting BLANC-help. The result is shown up in Figure \ref{fig:Summary_deterioration}.

\section{Comparison with human evaluation scores}
The BLANC measures do not require any "gold-labeled" data: No human-written summaries nor human-annotated quality scores are needed. Theoretically, the measures should reflect how fluent, informative, and factually correct a summary is, simply because only fluent, informative, correct summaries are helpful to the underlying language model. We now turn to the question of whether the BLANC measures correlate with summary quality scores assigned by human readers.

Human scoring is fallible; a correlation with human scores should not be considered as a full validation of our measures, but rather as an independent confirmation that the measures are sensible. 

For purposes unrelated to this study, we have undertaken a series of human evaluations of many generated summaries of approximately similar length. As mentioned in the previous section, the summaries were generated by Microsoft's abstractive UniML model \cite{Li2019Unified}, by semi-abstractive model \cite{oleg2019headline}, and by extractive LexRank model \cite{Gunes2004LexRank}. The summaries from the latter two sources were "equalized" in length to the UniML, so that at least on average the summaries from all three generation sources would be equal, and also so that most summaries would not differ significantly in length. Altogether, we assembled 555 summary-text pairs for human scoring, with the texts taken from the CNN / Daily Mail dataset \cite{Karl2015Teaching}.

We hired 10 annotators through Odetta.ai and trained them to assess the overall quality of each summary on a 5-point scale: 0 = VERY BAD, 1 = BAD, 2 = OK, 3 = GOOD or 4 = VERY GOOD. The annotators worked independently from each other and had access to only one summary-text pair at a time. The task was performed through the online text annotation tool LightTag (lighttag.io). 

\begin{figure}[htb]
    \centering
    \includegraphics[width=7.9cm]{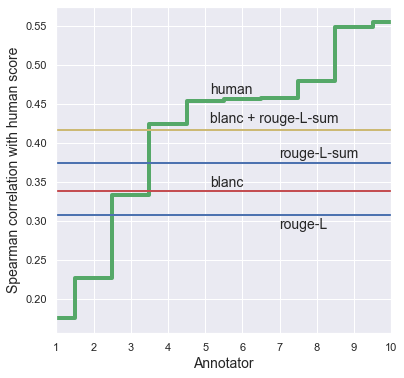}
    \caption{Spearman correlations with human annotators. The green step-line is the level of correlations of one of annotators with all other annotators. The correlation of BLANC-help with an average over all annotators is shown by the red line. The blue lines correspond to ROUGE-L and ROUGE-Lsum, and the yellow line to a simple sum "BLANC-help + ROUGE-Lsum". The summaries were generated on the CNN / DailyMail texts.}
    \label{fig:Corr_human_rouge}
\end{figure}

The values of the correlations are illustrated in Figure \ref{fig:Corr_human_rouge}.

The green step-function shows the value of correlation of an annotator score (with Id ranging from 1 to 10) with the averaged score of the 9 other annotators. The number of samples used for the correlation is 555 - the summaries generated by the three models. The red and blue lines show correlations of BLANC-help and rouge correspondingly with the averaged score of all 10 annotators. The rouge here is calculated using the google-research package (github.com/google-research/google-research/tree/master/rouge) as F1 value of "rougeL" (lower blue line on the plot) and F1 value of "rougeLsum" (upper blue line). The latter is the 'summary-level LCS', with summaries split to sentences and using a union longest common subsequence ~\cite{OriginalROUGE}

The yellow line in the figure shows how a simplest combination of BLANC-help and ROUGE correlates with the annotators. The "BLANC-help + ROUGE-Lsum" is literally a simple sum of BLANC-help and the ROUGE-Lsum. As usual a blending of two different models produces better results, though it is not our purpose here to fit human scores, and we do not fit the weights in the sum. (For example, using a $score = 3 * blanc_{help} + rouge_{Lsum}$ with the weight $3$ for BLANC-help would increase the correlation with human scores by 1\%).

All shown correlations have p-values of order $10^{-6}$ and lower. We observe that both BLANC-help and ROUGE correlate with annotators as good as or better than about 30\% of annotators. 

\begin{figure}[!htbp]
    \centering
    \includegraphics[width=7.9cm]{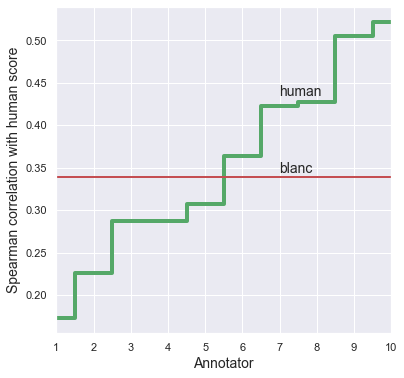}
    \caption{Spearman correlations with human annotators. The green step-line is the level of correlations of one of the annotators with all other annotators. The correlation of BLANC-help with an average over all annotators is shown by the red line. The summaries were generated on regular news documents: There are no reference summaries, and hence no ROUGE score.}
    \label{fig:Corr_human}
\end{figure}

In Figure \ref{fig:Corr_human} we present correlations with human scores on summaries generated for 100 typical daily news documents. The summaries were generated by the same three models; there were 300 summary-text pairs for scoring, again by 10 annotators. 

Since there are no "gold-labeled" summaries for these news documents, there is no ROUGE score in the figure.

As we see in all these examples, the human-human agreement is not impressive. We have observed from yet another evaluation dataset that if the texts and the generated summaries are challenging with very low inter-annotator agreement, the correlation of our measure with human scores is similarly diminished (with borderline p-values).

The values assigned to humans scores (0,1,2,3,4) are not a perfect translation of the human perception of the corresponding labels ("very bad", "bad", "OK", "good", "very good"). From multiple evaluations unrelated to this study we know that when an evaluation is repeated, human annotators are far more likely to substitute "OK" and "good" with each other than other tags. When we obtain an averaged human score, a weighting of the values $(0,1,2,3,4)$ with $weights=(3.0, 3.0, 1.0, 1.0, 2.0)$ may be more inline with human perception, but the changes to results we presented here are not essential, of order 1\%.

BERTScore \cite{Tianyi2020BERTScore}, similarly to ROUGE, requires reference summaries, but is using overlaps of BERT embeddings rather than strings. In Figure \ref{fig:Corr_human_rouge} the BERTScore F1 would be at 0.35 - close to BLANC, BERTScore Precision at 0.16, and BERTScore Recall at impressive 0.47 (calculated using python package bert-score).

A few simple observations may serve as an evidence that our measure deals with the length of a summary more reasonably than either humans or ROUGE. In Table \ref{tab:corr_len_summ_roughe} we show the correlations with the summary length and with the compression factor, which is defined as the ratio of summary length to document text length. The length here is the number of characters.

\begin{table}[htb]
  \begin{center}
  
    \begin{tabular}{l|l|r|r}
      \textbf{Estimator} & \textbf{Correlation} & \textbf{L} & \textbf{C}\\
      \hline
      {BLANC-help} & Pearson & 0.47 & 0.75\\
      & Spearman & 0.51 & 0.76\\ % <-- Content of first column omitted.
      \hline
      {rouge-L} & Pearson & -0.27 & \\
      & Spearman & -0.22 & \\ % <-- Content of first column omitted.
      \hline
      {rouge-L-sum} & Pearson & -0.23 & \\
      & Spearman & -0.15 & \\ % <-- Content of first column omitted.
      \hline
      {humans} & Pearson & 0.41 & 0.31\\
      & Spearman & 0.41 & 0.43\\ % <-- Content of first column omitted.
      \hline
    \end{tabular}
    \caption{Correlation of different quality estimators with length $L$ of summary and with compression $C$. The compression is defined as length of summary divided by length of text, in characters. The no correlation cases (p-value $> 0.05$) are left empty. Based on CNN / Daily Mail news.}
    \label{tab:corr_len_summ_roughe}
  \end{center}
\end{table} 

The table is based on the same data as Figure \ref{fig:Corr_human_rouge}. The table shows that similarly to humans, our measure is helped by longer summaries in general. But unlike humans, it is much more sensitive to a summary's compression factor. A disregard for the compression factor by humans may be caused by the anchoring effect.

Table \ref{tab:corr_len_summ} gives similar insight for very different kind of documents - random daily news, same as were used for Figure \ref{fig:Corr_human}.

\begin{table}[htb]
  \begin{center}
    \begin{tabular}{l|l|r|r}
      \textbf{Estimator} & \textbf{Correlation} & \textbf{L} & \textbf{C}\\
      \hline
      {BLANC-help} & Pearson & 0.20 & 0.77\\
      & Spearman & 0.19 & 0.73\\ % <-- Content of first column omitted.
      \hline
      humans & Pearson & 0.42 & 0.41\\
      & Spearman & 0.38 & 0.39\\ % <-- Content of first column omitted.
    \end{tabular}
  \caption{Correlation of different quality estimators with length $L$ of summary and with compression $C$. Based on randomly selected daily news documents.}
  \label{tab:corr_len_summ}
  \end{center}
\end{table} 

Whenever we used BLANC or human annotators for comparison of quality of summaries generated by different models of by different versions of a model, we generated summaries on average of the same length. It is clear that both humans and BLANC will estimate longer summary better, at least when it is a single score of overall summary quality. If the length of individual summary has to be excluded as a factor, the BLANC score should be normalized by the compression $C$. A longer summary adds proportionally more help, while a longer text adds proportionally more tokens for masking.

In Table \ref{tab:Jensen_Shannon} we show comparison of BLANC with negated Jensen-Shannon divergence (JS) which is a no-references measure showed up as the strongest in \cite{Louis2009Automatically}. The JS is a mean of text-summary and summary-text Kullback-Leibler divergences. For a purely statistical measure which we would assume misses a lot of semantics, JS works surprisingly well on CNN / Daily Mail news examples. The modest performance at first row by both measures can be explained by high variety of in styles of the summaries, which affects both the human scoring and the measures. On human-only summaries JS is still better than BLANC. In order to confirm that BLANC grasps more semantics, we considered three subsets of summaries that might have less signal from pure statistics. The summaries of similar length, close to peak of the distribution, is one example; summaries with low human scores is another one. More important example is the highly compressed summaries, with the ratio of the summary length to the text length $<0.05$. In this case JS correlation value would be $0.12$, but p-value=$0.15$ is too high. Following \cite{Louis2009Automatically}, the JS was calculated with filtering stop words and with stemming.

\begin{table}[htb]
  \begin{center}
    \begin{tabular}{l|r|r|r}
      \textbf{Selection} & \textbf{N} & \textbf{BLANC} & \textbf{JS}\\
      \hline
      All & 855 & 0.22 & 0.28\\
      \hline
      Human & 300 & 0.34 & 0.37\\
      \hline
      {Close length} & 319 & 0.15 & 0.13\\
      \hline
      Bad & 141 & 0.23 & 0.21\\
      \hline
      Compressed & 155 & 0.18 & (p=0.15)\\
    \end{tabular}
  \caption{Comparison of BLANC and Jensen-Shannon (JS) divergence correlations with averaged human score. First column specifies the summaries considered; second is the number of summaries; the last two columns are the correlations of BLANC and JS with human scores. The texts are from CNN / Daily Mail news. Row 'All' included all summaries, both human and generated by 3 methods. Row 'Human': only human-created summaries. Row 'Close length': summaries with length limited around pick of distribution, between 200 and 350 characters long. Row 'Bad' summaries with mean human score less than 2. Row 'Compressed': summaries with compression (length of summary over length of text) less than 0.05. There is no correlation in bottom JS cell, p-value=0.15.}
  \label{tab:Jensen_Shannon}
  \end{center}
\end{table} 

\begin{figure}[!htbp]
    \centering
    \includegraphics[width=8cm]{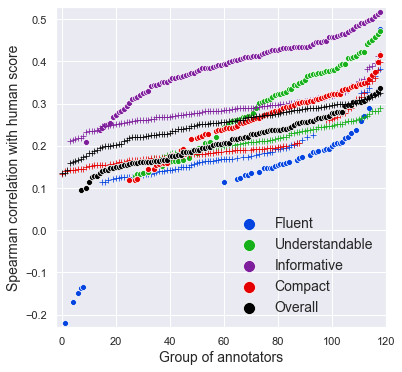}
    \caption{Spearman correlations with a group of 7 annotators. The x-axes depicts 120 ways to choose 3 annotators out of 10. The circle-markers show correlation of average score of 3 annotators with average score of 7 other annotators. The plus-markers show correlation of BLANC-help with the 7 annotators. Each type of correlation was sorted independently, left-to-right. Markers with p-values $> 0.05$ are not shown.}
    \label{fig:Corr_qualities}
\end{figure}

Simple correlation with a consensus score of annotators is not an easy criterion for judging the usefulness of the measure. When annotators are tasked with scoring several different qualities of a summary, their final score for the overall quality should be more grounded, because more attention has been spent on the summary and the text. In Figure \ref{fig:Corr_qualities} we show values of correlations obtained from such evaluation. 

The data used here are the same as the data for Figure \ref{fig:Corr_human}: summaries generated on randomly selected daily news documents. For this illustration, however, we split our 10 annotators into a small group of 3 and an "others" group of the remaining 7. There are 120 ways to chose the split (on the X-axis). The circle markers show human-human correlation, i.e. the correlation between the average score of the small group and the average score of the "others" group. The plus markers show BLANC-human correlation, i.e. a correlation of the BLANC with the "others" group of annotators. Hence we see how well the BLANC measure performs against the team of 3 annotators in correlating with the "others". For simplicity of the presentation, each type of correlation was sorted independently. If a correlation is unreliable (p-value $> 0.05$) then the marker is not shown.

We see that BLANC can be competitive to a team of three human annotators on all summary qualities, especially on the 'overall' and fluency.

\section{Conclusion}

In this paper we present BLANC, a new family of objective and reproducible measures of summary quality. BLANC does not require human-written reference summaries; it is based on how helpful the summary for understanding the document.  

By comparison, it is difficult to suspend disbelief when considering a method like ROUGE that does not inspect the document itself when estimating the quality of a summary. It is notable that ROUGE scores are often cited even for headline generation \cite{Ayana2016Neural, Shun2017Source, Peng2019Novel, Xiaotao2020Generating} where it is hard to imagine that any single headline could be regarded as the best possible headline for a document.

One may argue that ROUGE requires less processing power, unless we recall that applying it requires the processing power of a human who must write the reference summary for ROUGE. In future research we will consider variations of BLANC and, for convenience, provide a public package blanc.

We thank Charlene Chambliss (Primer) for help in preparing the design of human evaluations, and Rosanne Liu (UberAI), Nina Lopatina (In-Q-Tel) and anonymous reviewers for review of the paper and valuable feedback.

\bibliography{blanc}
\bibliographystyle{acl_natbib}

\end{document}